\documentclass[letterpaper, 10 pt, conference]{ieeeconf}  

\IEEEoverridecommandlockouts                              

\usepackage{adjustbox}
\usepackage[ruled, vlined]{algorithm2e}
\usepackage{amsmath}
\usepackage{amssymb}
\usepackage{array}  
\usepackage{authblk}
\usepackage[accsupp]{axessibility}  
\usepackage{booktabs}
\usepackage{capt-of}
\makeatletter
\let\NAT@parse\undefined
\makeatother
\usepackage{cite}
\usepackage[font={small}]{caption}
\usepackage{colortbl}
\usepackage{epsfig}
\usepackage{inconsolata}  
\usepackage{graphicx}
\usepackage{listings}
\usepackage{mathtools}
\usepackage{multicol}
\usepackage{multirow}
\usepackage{paralist}
\usepackage{pifont}
\usepackage{times}
\usepackage{tabularx}
\usepackage{xcolor}
\usepackage{svg}

\definecolor{flodarkpurple}{rgb}{0.288,0.1196,0.7}

\definecolor{amber}{rgb}{1.0, 0.75, 0.0}

\usepackage[backref=page,breaklinks,colorlinks,bookmarks,citecolor=flodarkpurple]{hyperref}

\newcommand{\webpage}{https://webpage.github.io/}

\newcommand{\authorhref}[3][flodarkpurple]{\href{#2}{\color{#1}{#3}}}

\title{\Large{\textbf{PickScan: Object discovery and reconstruction from handheld interactions}} \\
}

\author{
\authorhref{webpage}{Vincent van der Brugge}$^{1}$, 
\authorhref{webpage}{Marc Pollefeys}$^{1}$, 
\authorhref{webpage}{Joshua B. Tenenbaum}$^{2}$, \\
\authorhref{webpage}{Krishna Murthy Jatavallabhula}$^{2}$$^{\dagger}$, 
\authorhref{webpage}{Ayush Tewari}$^{2}$$^{\dagger}$ \\
\\
$^{1}$\href{affil}{ETH Zürich}, 
$^{2}$\href{affil}{MIT} 
$^\dagger$Equal advising
}

\begin{document}


\makeatletter
\let\@oldmaketitle\@maketitle
\renewcommand{\@maketitle}{\@oldmaketitle
\centering
\includegraphics[width=0.96\linewidth]{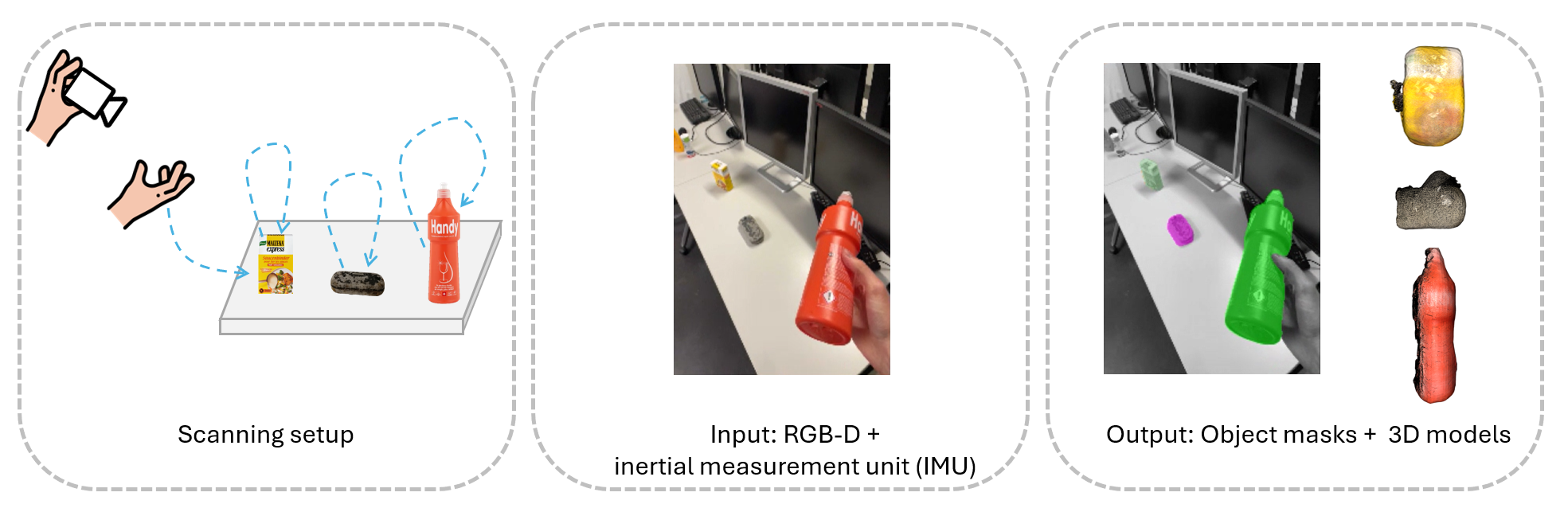}
\captionof{figure}{We present \textit{PickScan}, an interaction-guided and class-agnostic pipeline for compositional scene reconstruction. Our method lets a user pick up and move around objects, and outputs the object masks, 3D model and per-frame poses of each manipulated object.}
\label{fig:title_figure}
}
\makeatother

\maketitle
\thispagestyle{empty}
\pagestyle{empty}

\begin{abstract}

Reconstructing compositional 3D representations of scenes, where each object is represented with its own 3D model, is a highly desirable capability in robotics and augmented reality. However, most existing methods rely heavily on strong appearance priors for object discovery, therefore only working on those classes of objects on which the method has been trained, or do not allow for object manipulation, which is necessary to scan objects fully and to guide object discovery in challenging scenarios. We address these limitations with a novel interaction-guided and class-agnostic method based on object displacements that allows a user to move around a scene with an RGB-D camera, hold up objects, and finally outputs one 3D model per held-up object. Our main contribution to this end is a novel approach to detecting user-object interactions and extracting the masks of manipulated objects. On a custom-captured dataset, our pipeline discovers manipulated objects with 78.3\% precision at 100\% recall and reconstructs them with a mean chamfer distance of 0.90cm. Compared to Co-Fusion, the only comparable interaction-based and class-agnostic baseline, this corresponds to a reduction in chamfer distance of 73\% while detecting 99\% fewer false positives. 
\end{abstract}


\section*{Supplementary Material}
\textbf{Code:} \href{https://github.com/vincentvanderbrugge/pickandscan}{github.com/vincentvanderbrugge/pickandscan}

\setcounter{figure}{1} 

\section{Introduction}
\label{sec:intro}

Constructing detailed and informative maps of 3D environments is a fundamental task in robotics and augmented reality, and crucial for many downstream applications. A particularly important aspect of such maps is the degree to which they break down their surroundings into meaningful parts. For example, while many 3D reconstruction techniques reconstruct their surroundings as one ”contiguous” geometry, many applications would benefit from a breakdown into several manipulatable objects and a background that is considered to be static. Think of a robot finding and manipulating tools in an otherwise static environment, or an augmented reality experience where individual objects can be selected and hologram copies thereof edited, rearranged, etc.

In this work, we aim to design an easy-to-use pipeline that can scan environments in such a compositional way, resulting in reconstructions where different objects are represented with their own complete 3D model. We want said pipeline to work robustly on any type of object, as long as it can be grasped and held in a user's hand. Notably, we do not want to be restricted by the scanned objects having to fall within limited training distributions, as is typically the case when object discovery is driven by pre-trained segmentation networks.

\begin{figure}[!tbp]

\centering
  \vspace*{0.3cm}~\\
    \includegraphics[width=0.3\linewidth]{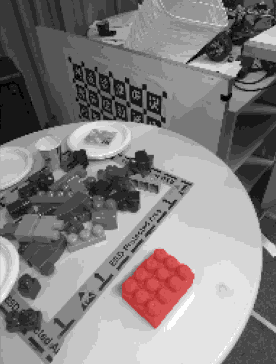}
  \hspace{0.1\linewidth}
    \includegraphics[width=0.3\linewidth]{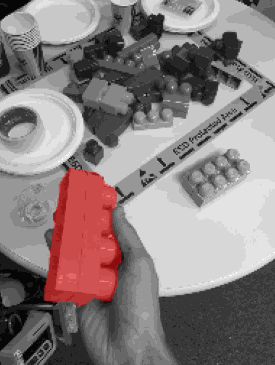}
  
  \caption{Example illustrating the benefit of user-object interactions for object discovery. Toy bricks of different sizes are placed next to each other in a 4 by 3 grid, out of which a 4 by 1 brick is picked up. \textbf{Left}: The closest mask found by a state-of-the-art segmentation network \cite{kirillov_segmentanything_2023} in the static scene in terms of intersection over union with the actual 4 by 1 brick's mask. \textbf{Right}: Object mask discovered by leveraging the handheld interaction, using our method.}
  \label{fig:lego_example_introduction}
  
\end{figure}

Nearly all existing methods to such compositional scene reconstruction fall short of these specifications. One critical aspect here is the way in which individual objects are identified and tracked. One prominent approach (cf. \cite{rosinol_kimera_2020}, \cite{mccormac_semantic_fusion_2016}, \cite{mccormac_fusionpp_2018},  \cite{tateno_seg_dense_slam_2015}) is to assume static scenes and identify objects using pre-trained segmentation networks. However, such approaches are inherently restricted to objects that fall within a limited training distribution. Moreover, the static assumption has the shortcoming that objects can not be manipulated to scan them fully from all sides, and provably makes object discovery impossible for some object constellations, for example for congregations of objects with similar texture but different shapes and sizes - see figure \ref{fig:lego_example_introduction} for an example. While a number of works have been proposed that allow for object movement in the scene (cf. \cite{barsan_cofusion_2018}, \cite{bescos_dynaslam2_2020}, \cite{xu_midfusion_2018}, \cite{ruenz_maskfusion_2018}), hence allowing for objects to be scanned from all sides, almost all of them still rely on pre-trained segmentation networks for object proposals, and are hence still restricted to fall within a limited training distribution of objects.

To our knowledge, the only existing method that fulfills our specifications, i.e. identifies objects in a class agnostic way while allowing for object movement to scan objects fully, is Co-Fusion \cite{chen_semantic_vslam_survey_2022}. It discovers objects by clustering points that move together in a rigid fashion, and is thus broadly applicable to any type of rigid object. However, to do this, Co-Fusion operates on the superpixel level, therefore limiting its accuracy, especially under sensor noise, and our evaluations show that the resulting discovered object masks are noisy, contain many false positive object detections, and the resulting object reconstructions tend to be of low quality.

In pursuit of an improved compositional scene reconstruction pipeline, we thus focus on object discovery and propose PickScan, our interaction-based, class-agnostic approach to object discovery and compositional scene reconstruction. Our approach operates on egocentric RGB-D streams of handheld user-object interactions, and additionally relies on inertial measurement unit (IMU) readings. We contribute a novel technique to detect the beginnings and ends of individual user-object detections and to propose one particularly representative mask for each manipulated object. Together with the state-of-the-art 2D mask tracker XMem \cite{cheng_xmem_2022}, the found object masks can be tracked to all other frames; and with the unknown object reconstruction technique BundleSDF \cite{wen_bundlesdf_2023}, the tracked object masks can then be used to reconstruct and track each discovered object in 3D, leaving us with a fully automated pipeline for compositional scene reconstruction. Note the difference between BundleSDF and our pipeline: While BundleSDF relies on object masks as an input, and can reconstruct the corresponding object from an RGB-D stream, our contribution lies in detecting one such mask for each manipulated object. This bridges the gap to an easy-to-use pipeline that can jointly discover and reconstruct multiple rigid objects in a complex scene without any prior class information. \\

To validate our method, we compare our method's performance with the state-of-the-art method Co-Fusion on a custom captured dataset. Moreover, we compare our object discovery method to the benchmark of using pre-trained segmentation networks. \\

In summary, our contributions are as follows:

\begin{itemize}
    \item A class-agnostic, interaction-based approach to discovering manipulated objects from egocentric RGB-D/IMU data.
    \item An architecture that combines this contribution with existing works to form a class-agnostic, interaction-based compositional scene reconstruction pipeline that outperforms the state-of-the-art.
\end{itemize}

\section{Related Work}
\label{sec:related_work}

Reconstructing 3D models of scenes has been widely studied, with a lot of early work done on structure from motion (SfM) and simultaneous localization and mapping (SLAM). While initial works focused on reconstructing a scene's geometry as a whole, some later works also break down the scene into individual objects. Many of these compositional scene reconstruction methods assume a scene to be static, i.e. do not allow for any object movement or dynamic interaction. Some notable SLAM-based examples include \cite{rosinol_kimera_2020}, \cite{mccormac_semantic_fusion_2016}, which reconstruct a static scene and break it down into semantic classes, and \cite{mccormac_fusionpp_2018},  \cite{tateno_seg_dense_slam_2015} which do the same on the instance-level. Such methods, which combine SLAM with some type of semantic information, belong to the category of semantic visual SLAM - see \cite{chen_semantic_vslam_survey_2022} for a comprehensive survey. Recently and outside the SLAM context, a number of neural-representation-based methods have also tackled static compositional scene reconstruction, see for example \cite{jatavallabhula_conceptfusion_2023}, \cite{vora_nesf_2021}.

More relevant to us are methods like \cite{barsan_cofusion_2018}, \cite{bescos_dynaslam2_2020}, \cite{xu_midfusion_2018}, \cite{ruenz_maskfusion_2018} that do not assume a scene to be static, but instead allow for objects to be moved around and hence scanned fully. Most of these methods still rely on semantic information, usually provided by a pre-trained segmentation network, to propose object instances, and thus are not class agnostic. The notable exception here is Co-Fusion \cite{barsan_cofusion_2018}, which can discover objects solely based on their rigid motion. To our knowledge, this is the only existing method that fulfills both the constraint of being class-agnostic as well as allowing for objects to be moved around by the user and hence scanned fully.

Lastly, two other lines of work are also relevant to our method: Firstly, handheld object reconstruction methods like \cite{wen_bundlesdf_2023}, \cite{huang_reconstructing_2022}, \cite{hampali_inhand_scanning_2023}, \cite{wang_demograsp_2021}, \cite{tzionas_3dhandobject_2015} deal with extracting 3D models of objects that are manipulated in a user's hand. However, these assume either a static camera or object masks to be given, and hence do not allow to reconstruct entire scenes in a compositional manner. Secondly, motion segmentation techniques like \cite{jain_accumdiff_1979}, \cite{hayman_statbackgroundsub_2003}, \cite{elhamifar_ssclustering_2012},  \cite{brox_longtermseg_2010}, \cite{liu_flownet3d_2019}, \cite{judd_multimotionvo_2018} identify individual objects based on them moving in a rigid fashion, although they do not reconstruct these objects. Related to this are methods like \cite{wu_d2nerf_2022}, \cite{tschernezki_neuraldiff_2021} which extend dynamic neural radiance fields \cite{pumarola_d-nerf_2020} to reconstruct dynamic scenes while separating them into moving and non-moving scene parts; these do not break down the scene into individual objects, however, and their performance is limited.

\section{Methodology}
\label{sec:methodology}

\begin{figure}
    \centering
    \vspace*{0.3cm}~\\
    \includegraphics[width=1.0\linewidth]{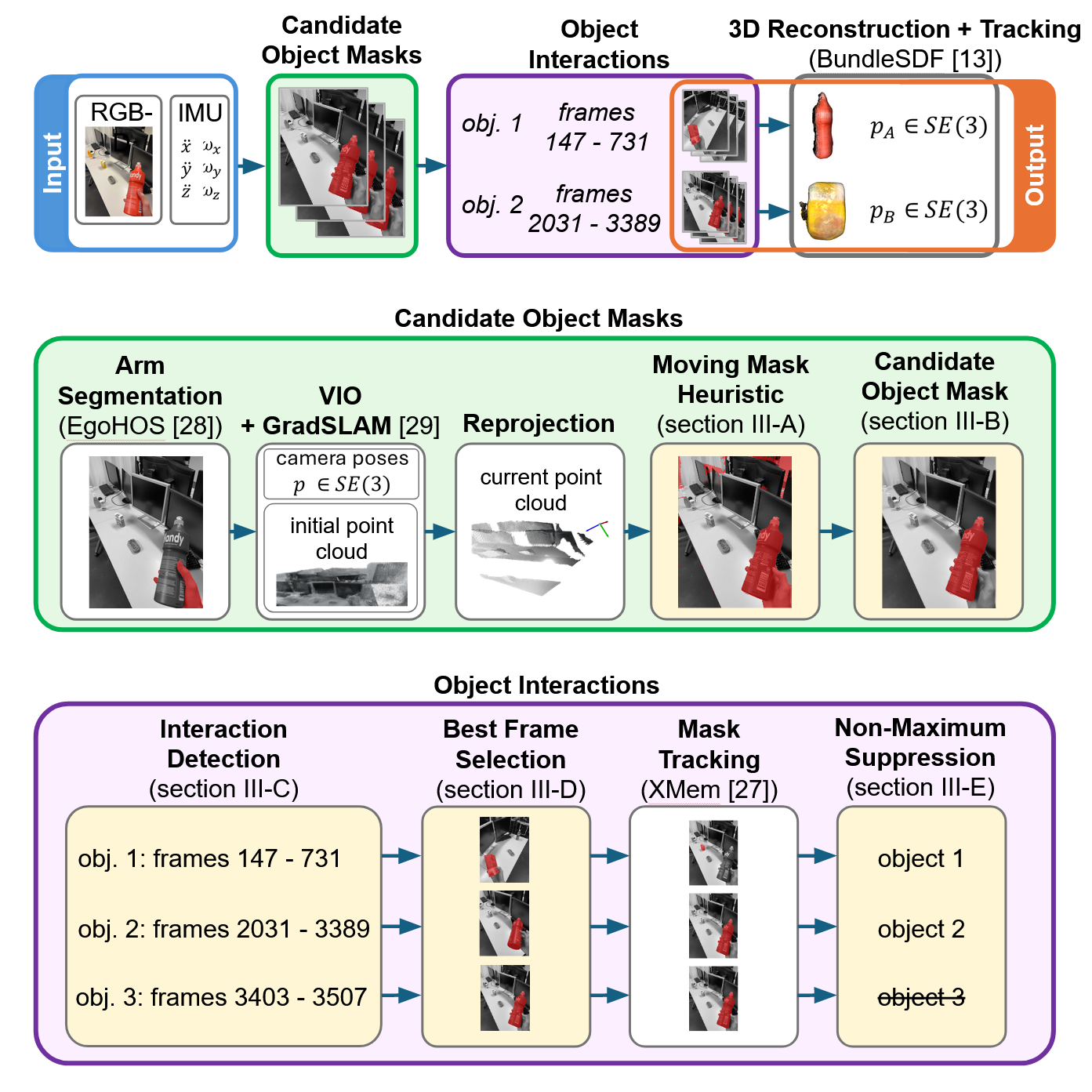}
    \caption{Overview of our pipeline and breakdown of its two phases: candidate object mask detection (\textcolor{green}{green}) and the object interaction phase (\textcolor{violet}{purple}). Our contributions are highlighted in yellow while existing components are shown with white filling.}
    \label{fig:method_overview}
\end{figure}

The main idea behind our pipeline is that, building on existing works, we can reduce the compositional scene reconstruction problem to finding the mask of each manipulated object in some frame of the input RGB-D stream. Using a 2D mask tracker like XMem \cite{cheng_xmem_2022}, we can then track this mask to all other frames; and with these propagated masks, we can reconstruct and track the respective object in 3D using an unknown object reconstruction and tracking technique like BundleSDF \cite{wen_bundlesdf_2023}.

The resulting pipeline is depicted in figure \ref{fig:method_overview} and can be thought to consist of two phases: In a first phase, we first estimate, for each frame, the mask of a potential manipulated object ("candidate object mask") - regardless of whether manipulation takes place in that frame. For this, we estimate a per-frame segmentation of the moving scene parts by comparing a given frame's reprojected point cloud with a point cloud reconstruction of the initial scene (section \ref{subsec:moving_mask}), which we then refine to per-frame candidate object masks (section \ref{subsec:candidate_object_mask}). In a second, "object interaction" phase, the candidate object masks are in turn used to estimate the masks for each respective manipulated object. Specifically, we first detect individual user-object interactions (section \ref{subsec:interaction_detection}). For each such interaction, we pick one particularly representative ("best") mask (section \ref{subsec:best_frame_selection}), track it in 2D, and suppress possible duplicate object detections (section \ref{subsec:nms}). After these two phases, we finally use the tracked masks to reconstruct and track the manipulated objects in 3D using BundleSDF \cite{wen_bundlesdf_2023}.

\subsection{Moving mask heuristic}
\label{subsec:moving_mask}

In our egocentric setup, the only possibly moving objects are the arm and potentially a manipulated object; our approach to estimating the manipulated object mask is thus to estimate the mask of all moving scene parts ("moving mask") in each frame before removing an estimated arm mask. To estimate the moving mask, we follow a displacement-based heuristic, where points are estimated to be moving if they are sufficiently far away from the initial scene at time $t = 0$. To reconstruct a point cloud of the initial scene ("initial point cloud"), we require the user to first scan the scene without manipulating any objects; after this static scan is completed, they can start manipulating and scanning objects. We use a pre-trained arm segmentation network \cite{zhang_egohos_2022} to identify the end of this static scanning phase, namely as the moment the arm mask stops being non-empty. Note that arm segmentation is the only task for which we use a pre-trained segmentation network; we do not use pre-trained segmentation networks to discover objects directly. The initial point cloud is extracted from the frames in the static scanning phase: Using initial pose estimates from a visual-inertial odometry (VIO) pipeline, we refine the camera poses and fuse the static phase's RGB-D frames into a point cloud with the differentiable SLAM pipeline gradSLAM \cite{jatavallabhula_gradslam_2019}. We also run this pipeline (VIO + gradSLAM) on \textit{all} frames of the scan (including ones where manipulation takes place) to extract all camera poses. With this, we can reproject a pixel $(u, v)$ at time $t$ into the canonical coordinate frame $F_0$ of the initial point cloud using perspective camera geometry:

$$\prescript{}{t}p_{(u,v)} = \pi^{-1}(u, v, d[t, u,v], K) = K^{-1} \cdot d[t, u,v] \cdot \begin{bmatrix} u \\ v \\ 1 \end{bmatrix}
$$

$$\begin{bmatrix} \prescript{}{0}p_{(u,v)} \\ 1 \end{bmatrix} = T_{0,t} \cdot \begin{bmatrix}
    \prescript{}{t}p_{(u,v)} \\ 1
\end{bmatrix}$$

Where $\prescript{}{i}p_{(u,v)}$ is the point's 3D coordinate in coordinate frame $F_i$, $K$ is the camera intrinsic matrix,  $d[t]$ the depth frame at time $t$, and $T_{0,t}$ the 4x4 homogeneous transformation from coordinate frame $F_t$ to $F_0$. We detect a pixel $(u, v)$ as moving if its reprojection is further away than a tunable threshold radius $\rho_{moving}$ from any point in the initial point cloud.

\subsection{Candidate object mask}
\label{subsec:candidate_object_mask}

Building on the estimated per-frame moving mask, we estimate a per-frame manipulated object mask next. For each frame, we remove the arm mask from the moving mask, divide the remainder into contiguous areas of pixels ("blobs"), and ignore blobs below a tunable threshold area in size. We then choose the blob with the reprojected points that are closest to the points in the user's hand. We estimate the hand mask as those pixels of the arm mask that occupy the topmost 10\% of its vertical pixel span.

\subsection{Interaction detection}
\label{subsec:interaction_detection}

\begin{figure}
    \centering
    \includegraphics[width=0.96\linewidth]{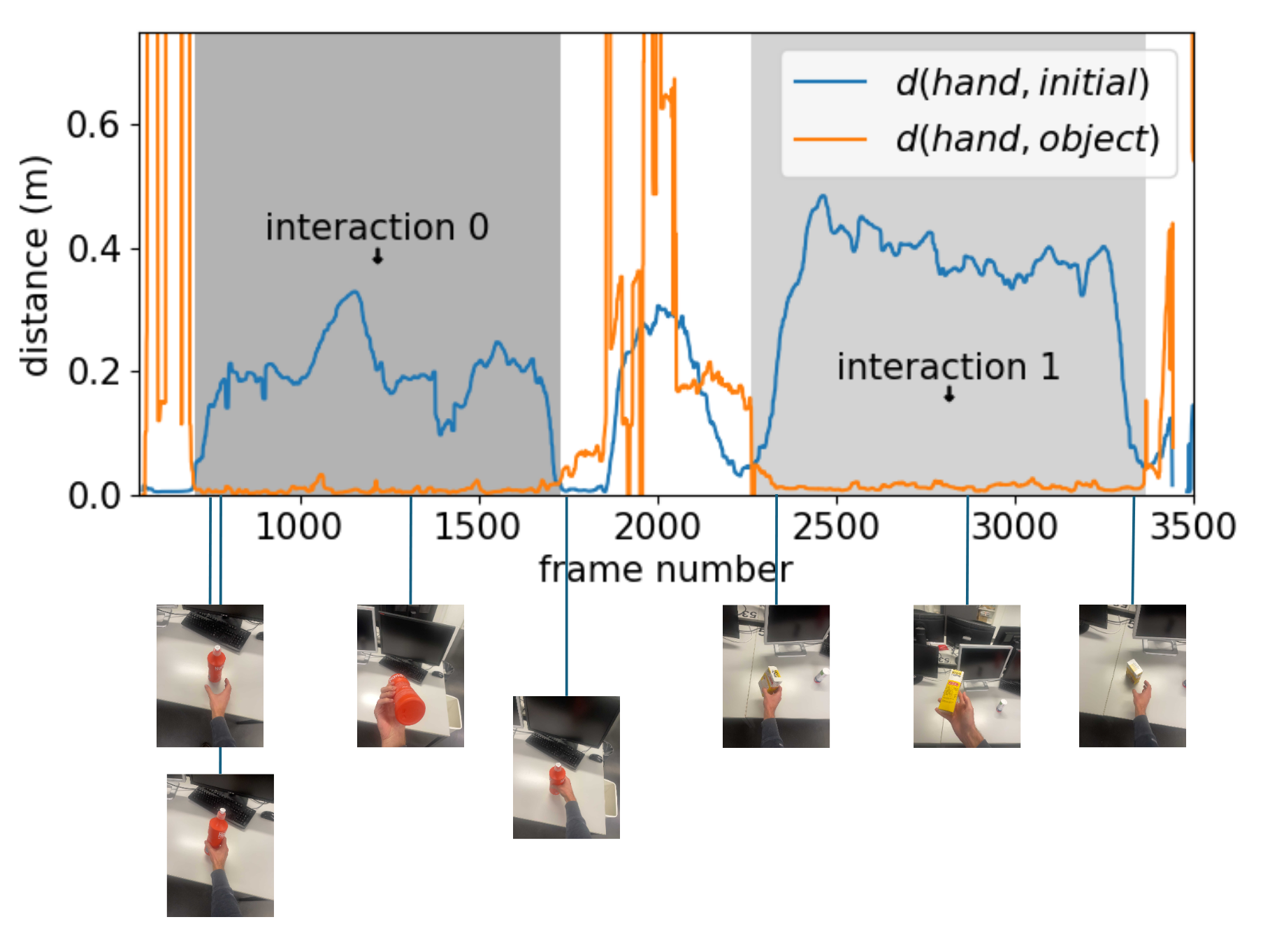}
    \caption{Interaction detection visualized: User-object interactions are detected as periods in which the distance between the hand point cloud and the initial point cloud (\textcolor{blue}{blue}) crosses over the distance between the hand point cloud and the candidate object point cloud (\textcolor{orange}{orange}), before crossing below it once again. In this example, detected interactions are highlighted in alternating shades of grey.}
    \label{fig:interaction-detection}
\end{figure}

To detect individual objects and hence factorize the scene, we detect the beginnings and ends of individual object interactions. For this, we focus on two different distances that evolve distinctively during a picking-up-and-laying-down maneuver. Firstly, the distance between the hand points and the initial point cloud ("hand-initial distance" hereinafter) is small as the interaction starts, becomes large as the user removes the object from its support (e.g. a table), and becomes small again as their hand once again approaches some support area while laying down the object. Meanwhile, the distance between the hand points and the candidate object mask's points ("hand-object distance" hereinafter) remains small during the entire interaction as the object is lodged in the user's hand. Following these patterns, we detect object interactions as periods starting with the hand-initial distance crossing above the hand-object distance, and ending with the former crossing below the latter once again. To minimize the effect of noise, we apply a median filter to both distance trajectories before extracting interactions. Lastly, we filter out interactions below a tunable threshold in duration for denoising. See figure \ref{fig:interaction-detection} for a visualization of this interaction detection step.

\subsection{Best frame selection}
\label{subsec:best_frame_selection}

Next, for each detected object (interaction), we pick a "best frame" in which the estimated object mask is deemed to be of high quality. For this, we employ the intersection-over-union (IoU) between subsequent frames' candidate object masks as a stability measure. Periods in which this cross-frame IoU remains above a tunable threshold $\tau_{iou}$ are deemed to be stable periods in which the mask is likely to be accurate, in contrast to periods with small cross-frame IoU in which the mask is changing drastically between frames. For the best frame, we focus on the longest stable period, and therein pick the frame with the highest cross-frame IoU with its previous object mask.


\subsection{Non-maximum suppression}
\label{subsec:nms}

After tracking each object's best frame with the mask tracker XMem \cite{cheng_xmem_2022}, we have a refined per-frame mask of each detected manipulated object. At this point, we remove duplicate object detections using a kind of non-maximum suppression, based on mask overlap: For each ordered pair (tuple) of detected objects (A, B), we calculate the mean percentage (over all frames) of object A's mask that is contained in object B's mask; we denote this (generally asymmetric) fraction as $contains_A(B)$. As long as a tuple of objects (A, B) exists for which $contains_A(B) > \tau_{nms}$, i.e. object A is contained to a sufficiently large extent in object B, we remove the object which is contained to the largest degree in some other object's mask, i.e. $\underset{A}{\arg\max} \ \underset{B}{\max} \ contains_A(B)$. At this point, the detected objects can be reconstructed and tracked using BundleSDF.

\\\\
\section{Experiments}
\label{sec:experiments}

To validate our method, we answer the following questions:

\begin{enumerate}
    \item How does our pipeline perform compared to Co-Fusion, the only baseline that is both class-agnostic and allows for scanning objects fully?
    \item Why use our interaction-based approach to object discovery rather than just finding objects using pre-trained segmentation networks?
\end{enumerate}

\subsection{Dataset}

\begin{figure}[h]

\centering
  \vspace*{0.3cm}~\\
    \includegraphics[width=0.25\linewidth]{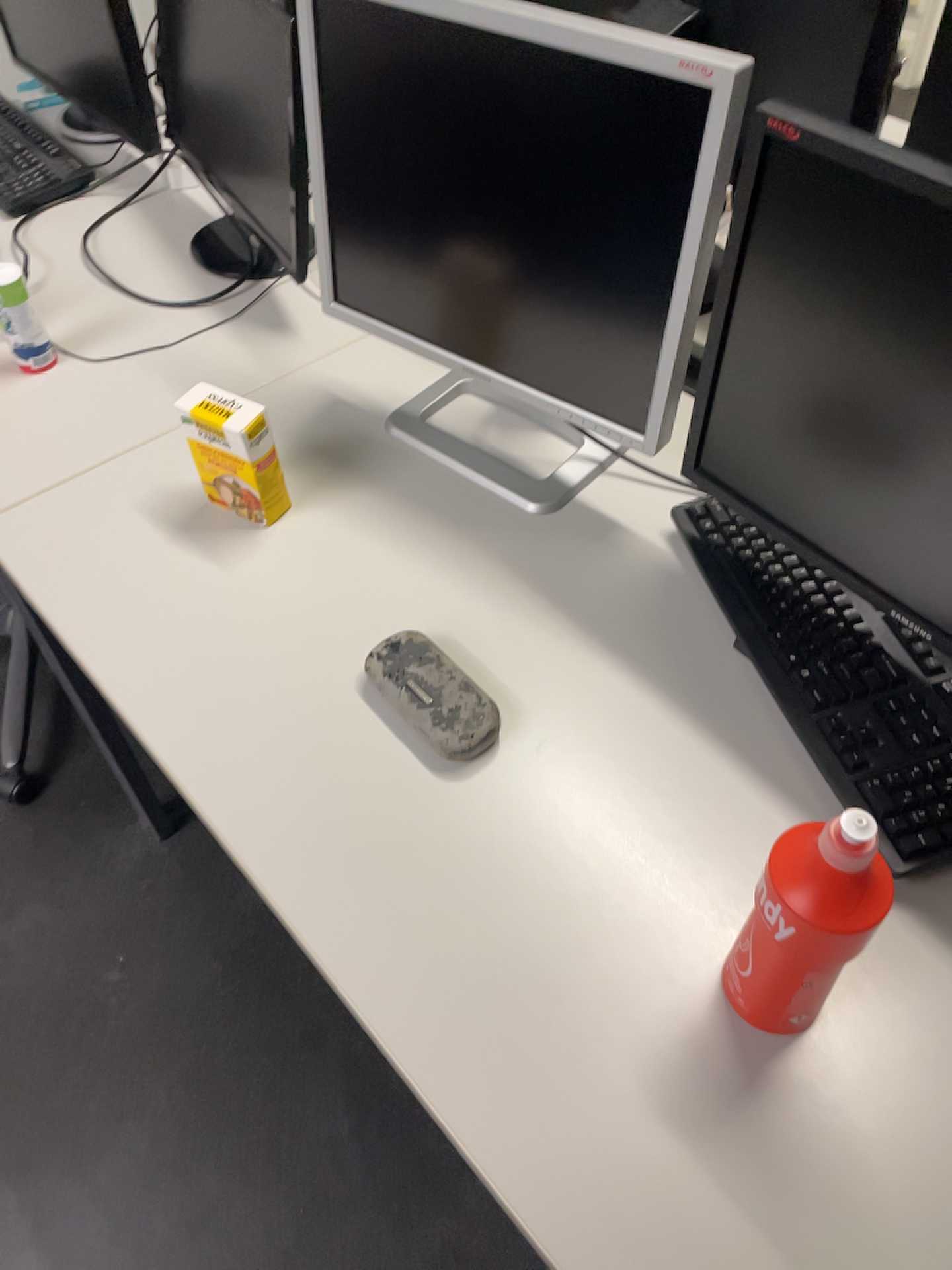}
  \hspace{0.05\linewidth}
    \includegraphics[width=0.25\linewidth]{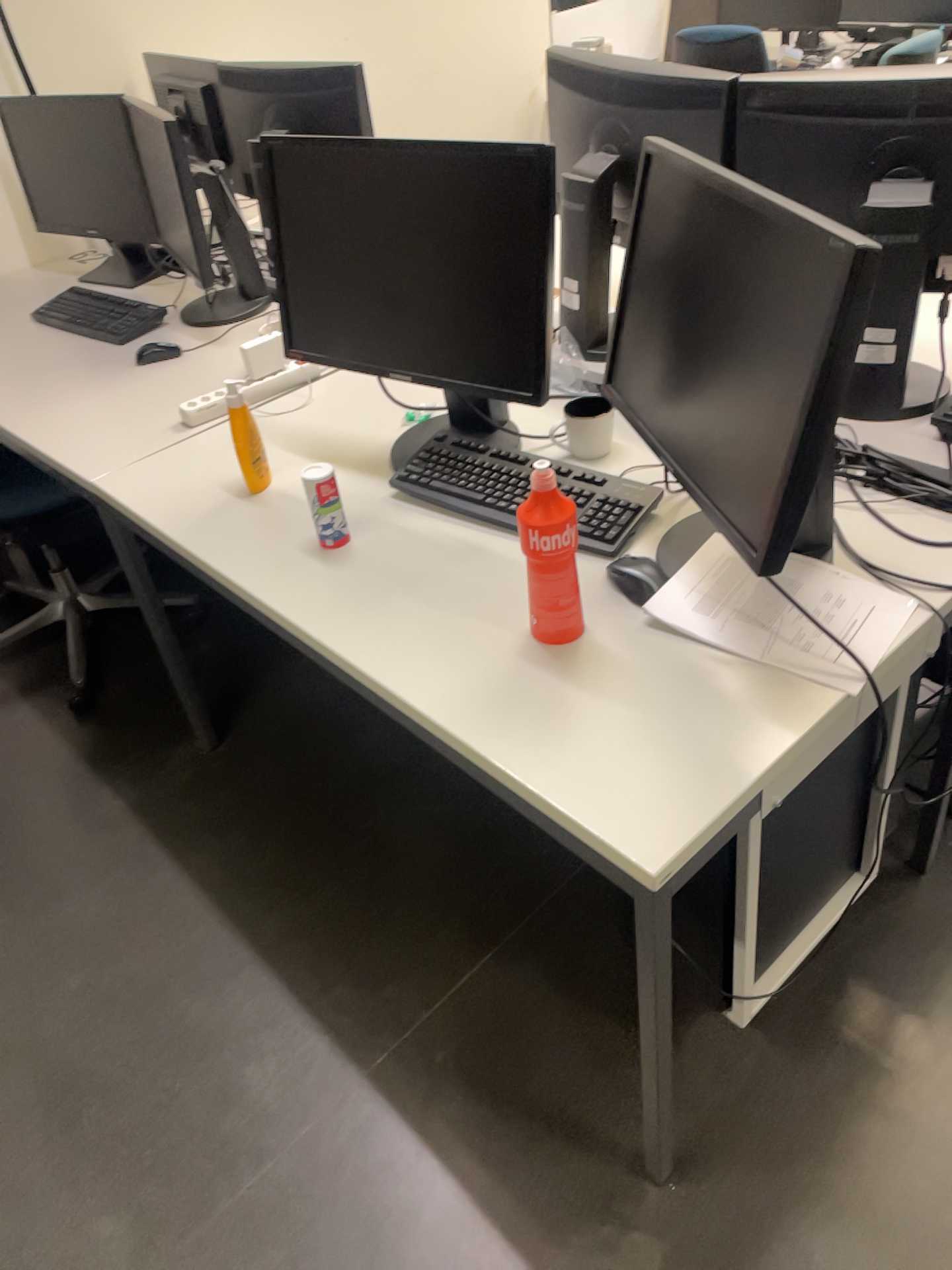}
  \hspace{0.05\linewidth}
    \includegraphics[width=0.25\linewidth]{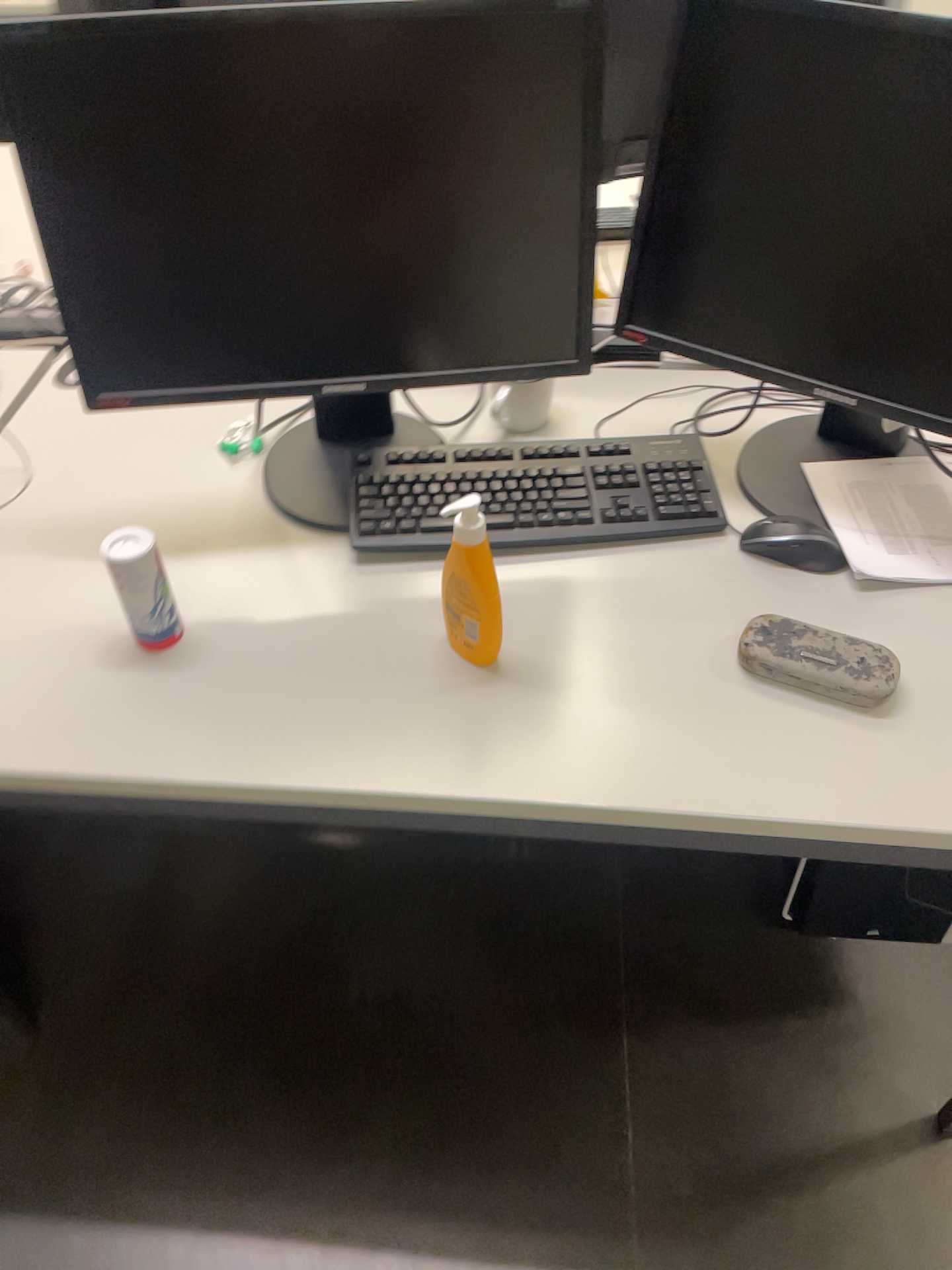}

  \caption{Scenes 1 to 3, from left to right, of our dataset. The images show the scenes in their initial state before any manipulation has taken place.}
  \label{fig:scenes_rgb}
  
\end{figure}

We evaluate on a custom dataset of 3 different scans in a tabletop setting, each containing 3 different objects that are manipulated and hence form the foreground under reconstruction (see figure \ref{fig:scenes_rgb}). The manipulated objects are taken from a set of 5 different foreground objects for which we have scanned high-fidelity 3D ground truth models with a static scanning setup \cite{einscan_3dscanner_2024}. We capture our scenes with an iPhone 12 Pro at a resolution of 192x256 pixels and a framerate of 60fps. A typical scan in our dataset lasts between 60 and 90 seconds, of which the first 10 to 30 seconds are made up of the static scanning phase for the reconstruction of the initial point cloud.

\subsection{Processing details}

We process our scans on a machine with an NVIDIA RTX A5000 GPU (24GB
GPU memory), 64GB RAM, and an Intel Core i9-10900X
CPU running at 3.70GHz. For dense reconstruction of a detected object with BundleSDF, we limit ourselves to processing every 5th frame in the span of the detected interaction to reduce the computational workload. With these settings, reconstructing one object with BundleSDF from a typical intermediary output of our pipeline (i.e. a stream of RGB-D frames and object masks of the detected object) with 700 frames, of which 140 are processed, takes around 60 minutes.

\subsection{Metrics}

We measure the chamfer distance between the output reconstructions and their respective ground truth models. Moreover, we count the number of detected false positive object detections in a scan and calculate the corresponding object detection precision and recall. For the calculation of the chamfer distance, we need to align reconstructions with their respective ground truth models. For this, we manually align the ground truth model with the reprojected point cloud in a suitable frame where the object is clearly visible and use BundleSDF's output pose in that frame to achieve a first alignment. After randomly sampling points from both meshes, we refine the alignment of the point clouds using the iterative closest point (ICP) algorithm before calculating the chamfer distance.

\subsection{Comparison with Co-Fusion}

\begin{figure}
    \centering
    \vspace*{0.1cm}~\\
    \includegraphics[width=0.96\linewidth]{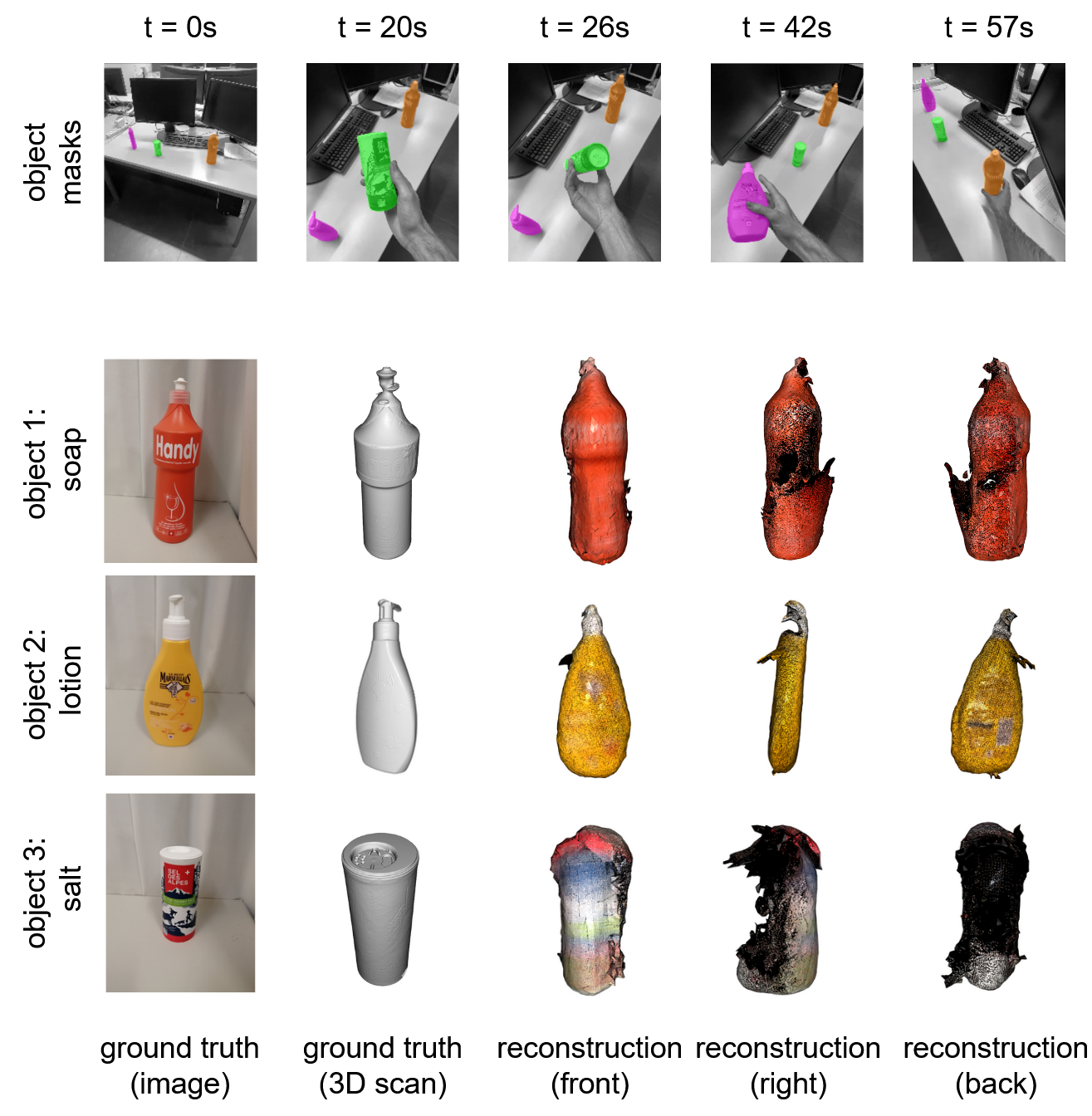}
    \caption{Output of our pipeline on an example scene from our dataset (scene 2). Discovered object masks are shown on the top, output object reconstructions and corresponding ground truth on the bottom.}
    \label{fig:example_results}
\end{figure}

\begin{figure}
    \centering
    \includegraphics[width=0.96\linewidth]{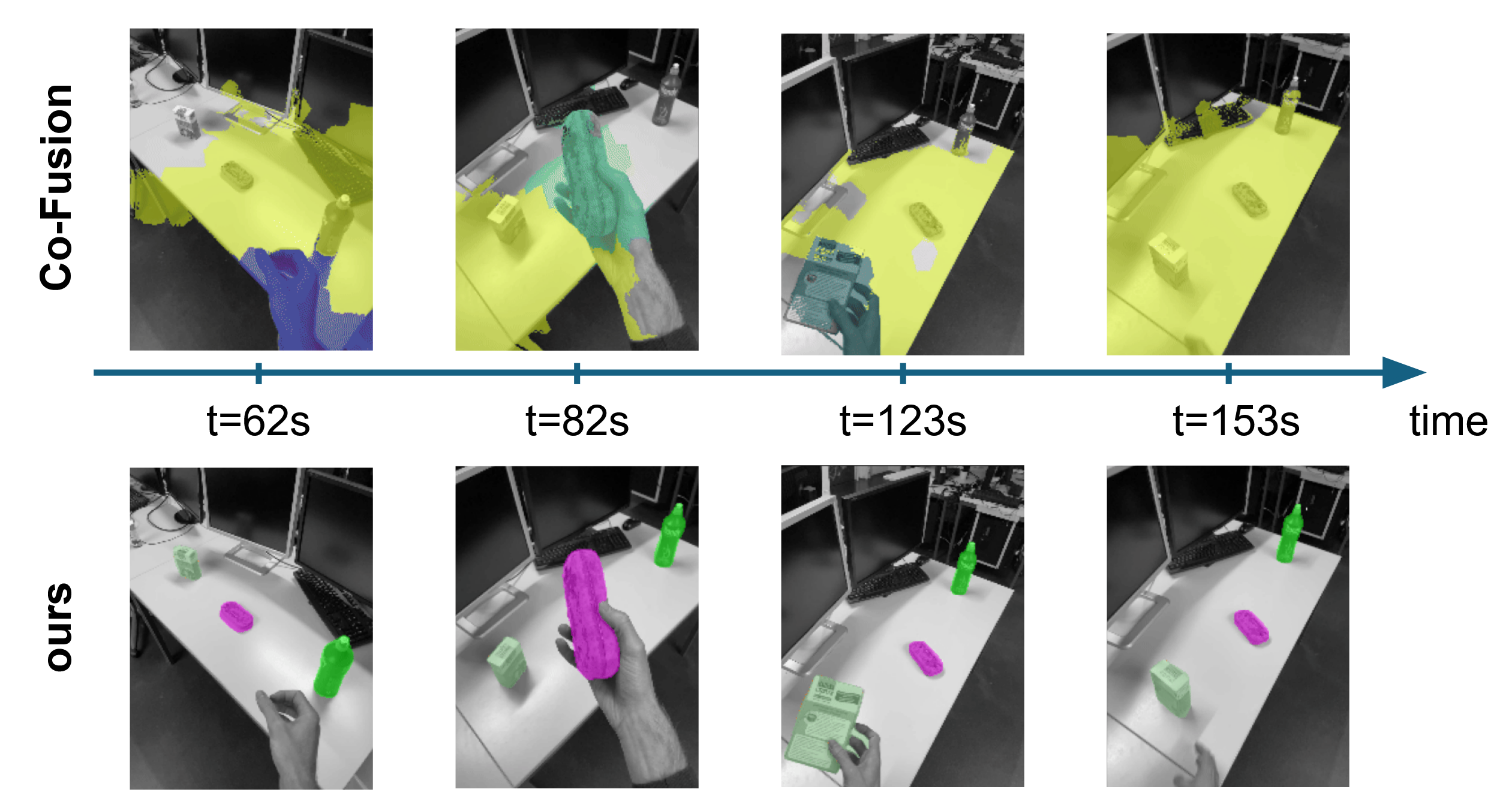}
    \caption{Qualitative comparison between the object masks provided by Co-Fusion versus ours. Our discovered object masks are generally of higher quality while containing less false positive object detections.}
    \label{fig:cofusion_versus_ours}
\end{figure}

View figure \ref{fig:example_results} for a qualitative inspection of our pipeline's output on one of our scenes, and table \ref{tab:results} for a quantitative performance summary. Our method outperforms Co-Fusion in every metric: It reconstructs objects with an average chamfer distance of 0.90cm versus Co-Fusion's 3.33cm while detecting only a fraction of its false positives. Consider figure \ref{fig:cofusion_versus_ours} for a qualitative comparison of Co-Fusion's object masks with those output by our system. Again, here it can be observed how Co-Fusion is more prone to false positive detections. Moreover, its discovered masks are significantly less accurate (cf. $t=82s$, $t=123s$), often do not correctly separate the manipulating hand (cf. $t=123s$), and, once having discovered an object, struggle to track it long-term (cf. $t=153s$).

\begin{table}[]

\begin{tabular}{@{}llllll@{}}
\vspace*{0.1cm}~\\
\toprule
  & \textbf{Method} & \textbf{Scene 1} & \textbf{Scene 2} & \textbf{Scene 3} & \textbf{Mean}   \\ \midrule
\multirow{3}{*}{\begin{tabular}[c]{@{}l@{}}mean chamfer\\ distance (cm)\end{tabular}} & ours      & \textbf{0.93}    & \textbf{0.64}    & 1.14             & \textbf{0.90}   \\
  & SAM*      & 1.16             & 1.37             & \textbf{0.98}    & 1.17            \\
  & Co-Fusion & 2.19             & 2.13             & 5.68             & 3.33            \\ \midrule
\multirow{3}{*}{false positives}                                                      & ours      & \textbf{0.0}       & \textbf{1.0}       & \textbf{2.0}       & \textbf{1.0}    \\
  & SAM*      & n/a              & n/a              & n/a              & n/a             \\
  & Co-Fusion & 63.0               & 126.0              & 113.0              & 100.7           \\ \midrule
\multirow{3}{*}{precision}                                                            & ours      & \textbf{100\%}   & \textbf{75\%}    & \textbf{60\%}    & \textbf{78.3\%} \\
  & SAM*      & n/a              & n/a              & n/a              & n/a             \\
  & Co-Fusion & 20.3\%           & 8.4\%            & 4.2\%            & 10.9\%          \\ \midrule
\multirow{3}{*}{recall}                                                               & ours      & \textbf{100\%}   & \textbf{100\%}   & \textbf{100\%}   & \textbf{100\%}  \\
  & SAM*      & n/a              & n/a              & n/a              & n/a             \\
  & Co-Fusion & \textbf{100\%}   & \textbf{100\%}   & \textbf{100\%}   & \textbf{100\%}  \\ \bottomrule
\end{tabular}
\caption{Evaluation results on our custom dataset. Our method significantly outperforms the state-of-the-art Co-Fusion, and yields similar results as when discovering objects using a state-of-the-art segmentation network (baseline SAM*), though our method has significant qualitative advantages over the latter (see section \ref{subsec:advantage_over_segmentation}).}
\label{tab:results}
\end{table}

What explains these differences in performance? Firstly, our method has a fundamentally different way of identifying manipulated objects: While Co-Fusion takes the more generally applicable approach of discovering objects as rigidly moving clusters of points, we rely on the picking-up-laying-down movement and identifying geometry that is far away from the initial scene, which we hypothesize to be a more robust clue for object discovery at the price of a more confined scanning process. Moreover, Co-Fusion operates at the superpixel level: after segmenting the image into superpixels, it proceeds with only one mean point per superpixel and performs rigid motion clustering on these center points. This arguably limits its accuracy, as manifested in the more noisy/coarse object masks.

\subsection{Advantage over segmentation-based object discovery}
\label{subsec:advantage_over_segmentation}

Given the prominence of semantic segmentation for object discovery tasks, we want to compare our method with a simple baseline where objects are discovered using semantic segmentation. For a given foreground object, we manually choose one frame in which the object is clearly visible, and manually annotate its ground truth mask in that frame. We run said RGB frame through a state-of-the-art segmentation network, specifically Segment Anything's (SAM) visual-transformer-based huge variant \cite{kirillov_segmentanything_2023}, and, of the resulting output masks, pick the one with the highest IoU with the object's ground truth mask. Leaving all other parts of the pipeline the same, we track this mask using XMem and reconstruct the object using BundleSDF. We denote this segmentation-based baseline SAM*. Note that the notion of false positives is meaningless for this baseline as we do not explicitly detect individual objects, but find a best-case segmentation mask for each object. Hence, the number of false positives and derived precision/recall measures are not reported for SAM*.

\begin{figure}
    \centering
    \vspace*{0.2cm}~\\
    \includegraphics[width=0.96\linewidth]{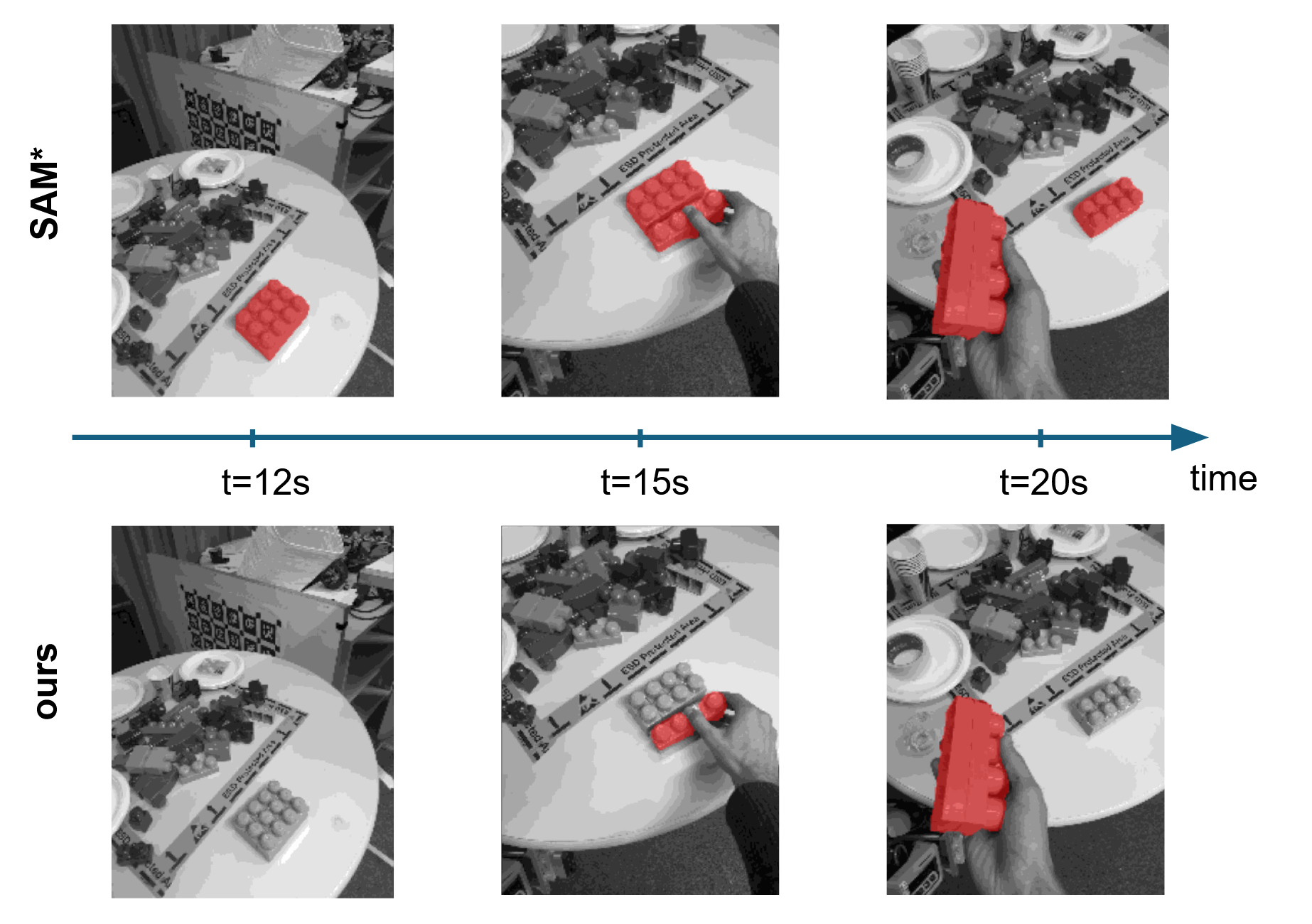}
    \caption{Visualizing the object discovery of our method versus that of a static segmentation-based approach (SAM*), using a toy brick example. Discovered \& tracked object masks are overlaid in red.}
    \label{fig:lego_example}
\end{figure}

Evaluated on our custom dataset, SAM* yields a mean chamfer distance of 1.17cm versus our 0.90cm, therefore demonstrating no downstream reconstruction advantage over our method of object discovery. However, there are important qualitative differences between our interaction-based approach to object discovery and this segmentation-based one. Firstly, our method is class-agnostic, while segmentation networks are inherently class-driven. Secondly, the fact that we can use interaction to guide object discovery makes it possible to correctly break down scenes with challenging object configurations, which is a particularly notable advantage over segmentation-based approaches that consider static scenes. Consider our experiment depicted in figure \ref{fig:lego_example}: Here, toy bricks of different sizes, including a 4 by 1 brick, are placed next to each other. Running SAM* on an image of the static scene ($t=12s$), we find that the brick is not correctly found by the segmentation network - the closest mask found is that of the entire brick grid. On the other hand, our method can correctly identify the 4 by 1 brick by leveraging the object interaction.

\subsection{Limitations}

We observe several limitations of our pipeline. Firstly, it still yields false positive object detections in some cases, which can generally be tracked to an inaccurate estimation of the distance between the hand point cloud and the initial point cloud, both of which exhibit some noise. Secondly, the mask tracking can fail in some cases, leading to object masks being lost, jumping to the wrong object at times, or to inaccurately covering the correct object. Lastly, as seen in figure \ref{fig:example_results}, the output reconstructions can display gaps, faulty geometry and/or faulty texture in some cases. Since this also occurs for objects for which the generated masks are quite accurate, we hypothesize that this is due to a failing on BundleSDF's part, particularly due to incorrect pose tracking. This explains why there are gaps despite the scans covering each object from all angles, and is in line with the fact that the object that consistently shows the highest reconstruction error (sunglasses case) is the object with the most symmetry and least texture, making it arguably the hardest to track. We also note that our input streams are of relatively small resolution, which likely adds difficulty to object tracking and reconstruction.

\section{Conclusion}

In this work, we have presented an interaction-guided, class-agnostic approach to compositional scene reconstruction. Specifically, our pipeline allows a user to move around a scene with an RGB-D camera, pick up and lay down objects, and outputs one 3D model for each picked-up object. Our main contribution is a mechanism to interaction-based object discovery that detects individual object interactions and outputs one mask for each manipulated object. Combined with existing works in neural-representation-based unknown object reconstruction as well as 2D mask tracking, this results in a fully automated, compositional scene reconstruction pipeline. In our evaluations, we have shown the significance of this pipeline: Firstly, we have shown that we outperform Co-Fusion, the only existing technique for compositional scene reconstruction that allows to scan objects fully while being class agnostic in its object discovery. Secondly, we have demonstrated our advantage over segmentation-based approaches, showing that our discovered object masks yield reconstructions of comparable quality downstream while having the advantage of being class-agnostic as well as being capable of leveraging object interactions to break down challenging scenes. \\

\noindent\textbf{Acknowledgements.}
This work was supported by the Swiss National Science Foundation under grant number 216260 (Beyond Frozen Worlds: Capturing Functional 3D Digital Twins from the Real World), the Singapore DSTA under DST00OECI20300823 (New Representations for Vision), and by the ONR MURI grant PO BB01540322.

\nocite{flaticon_camera_2024} \nocite{flaticon_hand_2024}



\bibliographystyle{IEEEtran}
\bibliography{IEEEtranBST/IEEEabrv, root}

\end{document}